\documentclass[final]{cvpr}

\usepackage{times}
\usepackage{epsfig}
\usepackage{graphicx}
\usepackage{amsmath}
\usepackage{amssymb}

\usepackage{array}
\usepackage{mathtools}
\usepackage{multirow}
\usepackage[english]{babel}
\usepackage[numbers,sort]{natbib} % sort citations
\usepackage{caption}
\usepackage{subfigure}
\usepackage{}\usepackage{algorithm}
\usepackage{algpseudocode}
\usepackage{makecell}
\usepackage{multirow}
\usepackage{booktabs}

\usepackage[pagebackref=true,breaklinks=true,letterpaper=true,colorlinks,bookmarks=false]{hyperref}

\usepackage{xspace}

\makeatletter
\DeclareRobustCommand\onedot{\futurelet\@let@token\@onedot}
\def\@onedot{\ifx\@let@token.\else.\null\fi\xspace}

\def\eg{\emph{e.g}\onedot} 
\def\ie{\emph{i.e}\onedot}

\def\etal{\emph{et al}\onedot}
\def\iid{\emph{i.i.d}\onedot}
\makeatother

\newcommand{\bowen}[1]{\textcolor{black}{#1}}

\newcommand{\algblue}[1]{\textcolor[rgb]{0.0, 0.0, 0.8}{#1}}
\newcommand{\algred}[1]{\textcolor[rgb]{0.8, 0.0, 0.0}{#1}}

\usepackage[pagebackref=true,breaklinks=true,colorlinks,bookmarks=false]{hyperref}

\def\cvprPaperID{****} % *** Enter the CVPR Paper ID here
\def\confYear{CVPR 2021}

\begin{document}

%%%%%%%%% TITLE
\title{A Simple Non-\iid Sampling Approach for Efficient Training and\\Better Generalization}

\author{
  %\begin{tabular}[t]{c}
Bowen Cheng$^{1}$, Yunchao Wei$^{1}$, Jiahui Yu$^{1}$, Shiyu Chang$^{2}$, Jinjun Xiong$^{2}$,\\Wen-Mei Hwu$^{1}$, Thomas S. Huang$^{1}$, Humphrey Shi$^{3,1}$\\
\\
{$^1$UIUC \hspace{2mm} $^2$IBM Research \hspace{2mm} $^3$U of Oregon}
}

% \author{First Author\\
% Institution1\\
% Institution1 address\\
% {\tt\small firstauthor@i1.org}
% % For a paper whose authors are all at the same institution,
% % omit the following lines up until the closing ``}''.
% % Additional authors and addresses can be added with ``\and'',
% % just like the second author.
% % To save space, use either the email address or home page, not both
% \and
% Second Author\\
% Institution2\\
% First line of institution2 address\\
% {\tt\small secondauthor@i2.org}
% }

\maketitle

%%%%%%%%% ABSTRACT
\begin{abstract}
% \textcolor{red}{Bowen: make table 4 larger and try to polish again from the intro to experiments. Also please format figures and tables. I will start to polish the paper wholly once these are done.}

While training on samples drawn from independent and identical distribution has been a \emph{de facto} paradigm for optimizing image classification networks, humans learn new concepts in an easy-to-hard manner and on the selected examples progressively. Driven by this fact, we investigate the training paradigms where the samples are not drawn from independent and identical distribution. We propose a data sampling strategy, named Drop-and-Refresh (DaR), motivated by the learning behaviors of humans that selectively drop easy samples and refresh them only periodically. We show in our experiments that the proposed DaR strategy can maintain (and in many cases improve) the predictive accuracy even when the training cost is reduced by 15\% on various datasets (CIFAR 10, CIFAR 100 and ImageNet) and with different backbone architectures (ResNets, DenseNets and MobileNets). Furthermore and perhaps more importantly, we find the ImageNet pre-trained models using our DaR sampling strategy achieves better transferability for the downstream tasks including object detection (\(+0.3\) AP), instance segmentation (\(+0.3\) AP), scene parsing  (\(+0.5\) mIoU) and human pose estimation (\(+0.6\) AP). Our investigation encourages people to rethink the connections between the sampling strategy for training and the transferability of its learned features for pre-training ImageNet models.

\end{abstract}

%%%%%%%%% BODY TEXT
\section{Introduction}

Human beings are exposed to a world with data at a much larger scale even compared with the largest datasets but they can still learn well efficiently. We observe that when human beings begin to learn and develop their intelligence, they (common people) tend to start with learning a broad area of knowledge without digging into the depth (similar to using all training examples). As the learning process proceeds, people will focus on a narrow and much more difficult area (a ``drop'' in the number of training examples). When people start to forget what they have learned before, however, people can quickly catch up by refreshing all materials in a short time (a ``refresh'' in easy examples). Furthermore, we notice that training samples in human learning process are not \iid. Easier samples are refreshed less frequently than harder samples. And the definition of easy and hard samples also change overtime, depending on the order of samples presented to human learner (\eg learning physics without math background makes all physics concept ``hard'' samples to learn). Motivated by this observation of human learning process, we investigate an efficient training method named Drop-and-Refresh (DaR) to imitate this process, so that all the training data can be sample in a more efficient way and not necessarily be \iid anymore.

\begin{figure*}[ht!]
	\centering
	\includegraphics[width=1.0\linewidth]{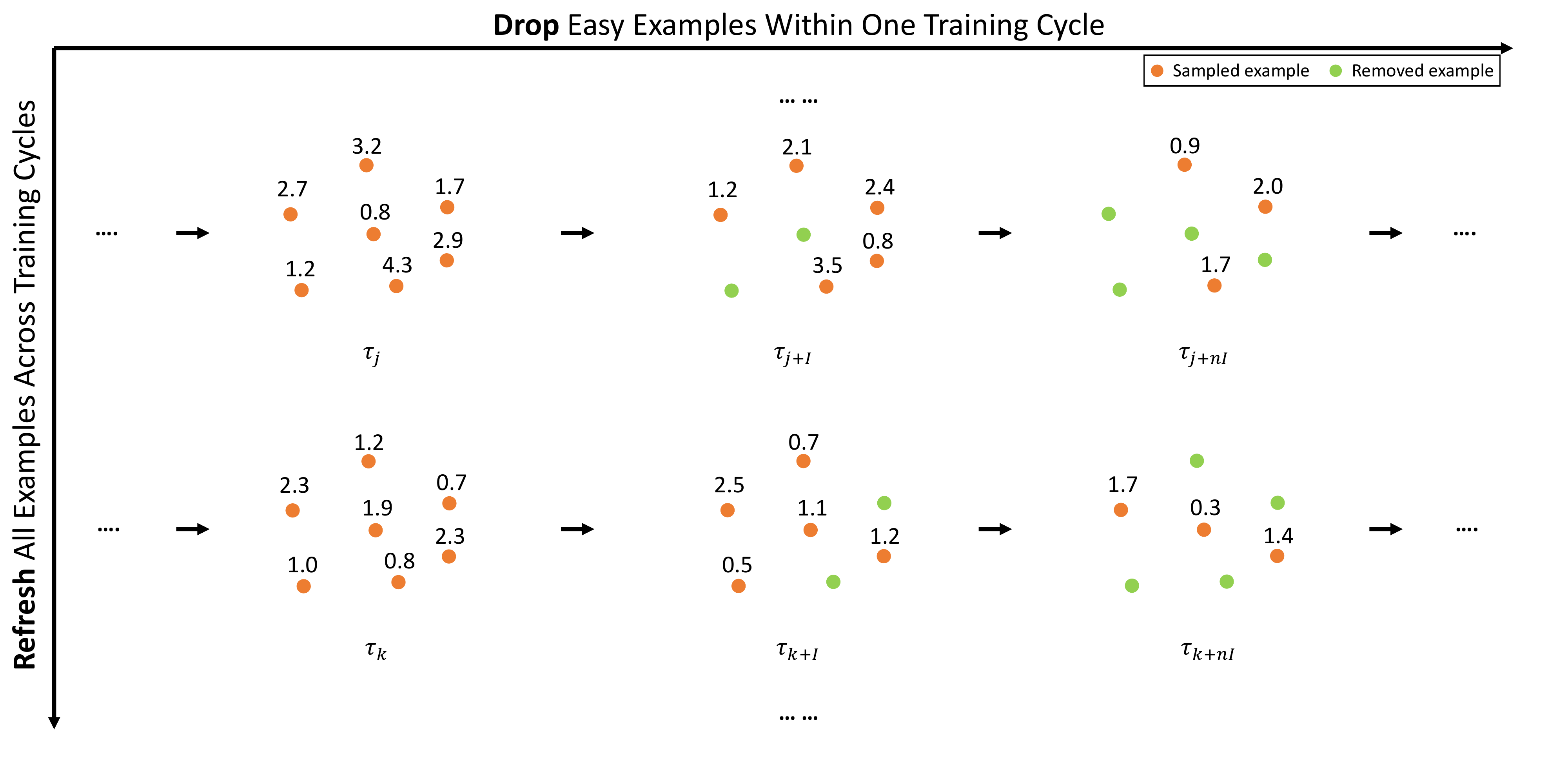}
	\caption{An illustration of Drop-and-Refresh (DaR). First, all the available training examples (orange dots) are adopted for training at the beginning of the training cycle \emph{j}. Second, as the training goes on, we gradually discard training examples (gray) to save training computation. Third, all the discarded samples are refreshed for training when a new cycle \emph{k} is started. We periodically repeat such a drop-and-refresh training manner until convergence. $\tau$ denotes a training epoch and $\mathcal{I}$ denotes interval. \textbf{Numbers above each sample point denotes the loss of that sample at current epoch.} Samples are removed by dropping examples with lowest losses and the discarded samples might be different in different training cycles.}
	\label{fig:training_policy}
\end{figure*}

Image classification is one of the most fundamental problem in computer vision. Pre-training on a large-scale image classification dataset has shown to be effectiveness for improving the performance of many downstream visual recognition tasks, \emph{e.g.} object detection \cite{ren2015faster,lin2017feature,he2017mask,lin2017focal}, semantic segmentation \cite{zhao2017pyramid,deeplabv3plus2018}, human pose estimation \cite{xiao2018simple,sun2017integral} \emph{etc.} Although efforts have been made to find more advanced architectures \cite{Shen2017DSOD,shen2017learning} or normalization method \cite{wu2018group} to train these downstream tasks from scratch without pre-training \cite{he2018rethinking}, the gap (either performance~\cite{Shen2017DSOD} or computation~\cite{he2018rethinking}) is still not negligible. The ImageNet \cite{deng2009imagenet} has been used as the standard pre-training dataset for a long time.  Many recent works \cite{deeplabv3plus2018,mahajan2018exploring} find that the larger dataset used for pre-training, the larger gain downstream tasks will benefit from. Since people usually follow the \iid assumption to uniformly sample training data, the computation is proportional to the size of the dataset. As the larger dataset is considered, the process of pre-training brings new challenges in terms of both of the budget (resources) and time.  For example, the pre-trained model in \cite{mahajan2018exploring} is trained on 336 GPUs across 42 machines for 22 days of training. Both of the training time and computational resources are likely not affordable for many research groups.  Thus, it would be helpful if one can find a more efficient training procedure to reduce the training time, which makes the pre-training can better serve for other downstream tasks. \bowen{Beyond the complexity of the network structure, the increase in pre-training time is due to the increase in the number of training examples. In general, current dataset construction techniques simply sample images randomly from the web (\emph{e.g.} Flickr, Instagram) with some keywords or hashtags \cite{deng2009imagenet,mahajan2018exploring}. However, the contribution of each example varies significantly during training, which has not been taken into account by current training manner. We ask the question that can we come up with a more efficient training strategy by taking into account the contribution of each example to \textbf{reduce the time of pre-training without affecting the generalization ability of the learned parameters for the downstream visual recognition tasks}?}

The basic idea of DaR is to divide the training process into different cycles. In each cycle, the network first learns from all training examples (this is similar to the stage when human learns a broad knowledge). As the training process goes on, we gradually feed the network with only a subset of the training examples (similar to human learning a specific subject). To prevent the model from forgetting what it learned before, we periodically repeat this training cycle where all training data is used again and restart the process of reducing training data for each training epoch. We perform extensive experiments on a variety of benchmarks (CIFAR10, CIFAR100, ImageNet) for image classification to first verify the efficiency and effectiveness of our proposed method.  Specifically, our proposed method only requires 83\%, 92\%, and 86\% training computation to achieve comparable or better results (\emph{w.r.t.} using all training examples) on CIFAR10, CIFAR100, and ImageNet respectively. Furthermore, when using DaR pre-trained ImageNet models to finetune on various downstream tasks, including \textbf{object detection, instance segmentation, scene parsing and human pose estimation}, we find these models achieve better performance than models pre-trained with \iid ImageNet training data in all tasks compared.

Our contributions are threefold:
\begin{itemize}
    \item 
    We present an efficient sampling method named Drop-and-Refresh (DaR) to sample a subset of non-\iid training data that can save training computation while retaining the performance.
    \item
    We validate the effectiveness of the DaR strategy on various benchmarks (\ie CIFAR10, CIFAR100 and ImageNet) with different backbone settings (\ie ResNet, DenseNet and MobileNet). DaR strategy can preserve the accuracy while reducing 15\% training cost.
    \item
    We further show that DaR pre-trained models have better generalization when finetuning on four different downstream tasks and we give a hypothesis to explain why models trained with non-\iid data generalize better.
\end{itemize}

\section{Related Works}
% The propose of our method is to accelerate the speed in the training stage. In this section, we compare different methods that reduce the training time.

\subsection{ImageNet Pre-training}
Recently, He \etal~\cite{he2018rethinking} find that the performance of training object detection and instance segmentation models from random initialization is no worse than their ImageNet pre-training counterparts on the COCO dataset~\cite{lin2014microsoft} without changing hyperparameters or model structures~\cite{Shen2017DSOD,zhu2018scratchdet,shen2019improving}. However, the conclusion in~\cite{he2018rethinking} only holds under certain circumstances.

The first requirement is that there is enough computational resource. To match the performance of ImageNet pre-training models, training from random initialization needs to be trained for $5\sim6$ times more iterations. Since most of time we fix the backbone network in object detection, it is a waste of computation to train object detection models with the same backbone. Another requirement is that there is enough target data which is hardly satisfied in all domains.

In our work, we find that when sampling non-\iid training data in a specific pattern, we can train image classification models more efficiently with less computation and still match the performance of using all training data. Furthermore, we find non-\iid pre-trained models generalize better to downstream tasks than pre-trained models using all training data.

\subsection{Fast Training Methods}
Works have been done to explore methods for training a model as fast as possible assuming \textbf{unlimited} computational resource. Large batch size is used in \cite{goyal2017accurate} with a carefully designed learning rate warm-up schedule. \cite{goyal2017accurate} trains a ResNet-50 model in 1 hour with 256 GPUs without loss in accuracy.  Mixed precision training is explored in \cite{micikevicius2017mixed} where the forward pass and backward pass are computed in half precision and parameter update in single precision. With the latest tensor core in the Nvidia Volta architecture, mixed precision can speed-up training by nearly 50\%. A combination with faster inter-node communication has been studied in \cite{sergeev2018horovod,jia2018highly}. In particular, \cite{jia2018highly} reports the training time of a ResNet-50 model is 7 minutes with 1024 GPUs without loss in accuracy.  Although fast training has been studied, the amount of computational resource is intractable for small research labs or individual researchers. In this paper, we try to speed-up training with \textbf{limited} computational resource (\emph{e.g.} with 4 GPUs available). And our method is in orthogonal with \cite{goyal2017accurate,micikevicius2017mixed,sergeev2018horovod,jia2018highly} and we believe the proposed method can further speed-up training of these methods.

% \subsection{Efficient Network Design}
% Designing efficient networks with more efficient operators is yet another way to reduce the training time. The MobileNet \cite{howard2017mobilenets,sandler2018mobilenetv2} family replace a regular $3\times3$ convolution by a $3\times3$ depth-wise separable convolution which is consist of a $3\times3$ depth-wise convolution and a regular $1\times1$ convolution. The ShuffleNet \cite{zhang2017shufflenet,ma2018shufflenetv2} family further replace the regular $1\times1$ convolution by a $1\times1$ group convolution and introduce a channel shuffle operation to break the channel dependency.  However, one problem of the efficient network is that it usually reduces the computation in the cost of loss in accuracy. Yet another problem is that experiments show the convergence of efficient networks is slower. A ResNet \cite{he2016deep} converges normally in 90-120 epochs, however, it takes more than 480 epochs to train smaller networks \cite{sandler2018mobilenetv2,zhang2017shufflenet}. Thus, an efficient network design might not always result in faster training time. 
% \tiger{I think related work also need to mention "active learning", "curriculum learning", and maybe "Pre-training".}

\subsection{Active Learning}
\bowen{Active learning is a method to expand the size of datasets by selecting examples that are most helpful for a specific network to label. The key idea in active learning is to select the most uncertain \textbf{unlabeled examples} to label. In some sense, this idea is relevant to our proposed strategy that we keep the most uncertain \textbf{labeled examples}.}

\bowen Joshi \etal {\cite{joshi2009multi} proposes to use an SVM classifier where uncertainties are calculated based on probabilistic outputs over the class label. Osband \etal \cite{osband2016deep} shows how to obtain uncertainty measures with neural networks. Lakshminarayanan \etal \cite{lakshminarayanan2017simple} uses extra head in neural network that is trained to estimate the variance. It allows variance along with predictions to be used to estimate uncertainty. In this work, we simply use the training loss as an indicator of the uncertainty of an example. The larger the loss, the more uncertain the example is to the network.}

\subsection{Hard Example Mining}
Hard example mining has been widely studied in the case of classical support vector machines~\cite{felzenszwalb2009object,ogawa2014safe} which find small number of hard negative samples from a large pool of negative samples. Recent works use importance sampling scheme to find most ``informative'' examples~\cite{katharopoulos2018not,schaul2015prioritized,loshchilov2015online} to speed-up the convergence of stochastic gradient descent. Online hard example mining (OHEM)~\cite{shrivastava2016training} is also commonly used in training object detection models, however, OHEM does not save any computation cost. Different from previous hard example mining approaches, our work not only focuses on efficient training but also studies the generalization of models trained with hard examples.

\section{Approach}
\setlength{\tabcolsep}{2pt}
\begin{table}[t!]
\begin{center}

\begin{tabular}{l|l}
% \hline\noalign{\smallskip}
\toprule
$\mathcal{X}$, $\mathcal{Y}$ & set of all training images, labels\\
$\mathcal{X}_{\text{s}}$, $\mathcal{Y}_{\text{s}}$ & set of sampled training images, labels\\
% $\mathcal{Y}$ & set of all training labels\\
$\mathcal{L}$ & set of training losses\\
$\mathcal{O}$ & set of network outputs\\
% $\mathcal{Y}_{\text{s}}$ & set of sampled training labels\\
$b$ & sampled mini-batch\\
$\mathcal{E}$ & number of training epochs\\
% \hline
% $\mathcal{W}$ & \makecell[l]{number of warm-up epochs\\where no subsampling occurs}\\
% \hline
% $\mathcal{I}$ & \makecell[l]{number of interval epochs\\between two subsamplings}\\
% \hline
% $p$ & \makecell[l]{percentage of samples to keep\\when subsampling (\emph{w.r.t.} the\\current number of examples)}\\
% \hline
% $\alpha$ & \makecell[l]{number of active epochs\\that subsampling is allowed\\every time all data is used}\\
% \hline
% forward(X) & forward a batch\\
% loss(X, Y) & calculate the loss\\
% backward(X) & backward loss, accumulate gradient\\
% sort(X) & function to sort elements in X\\
% \hline
% sample(X, $p$, Y) & \makecell[l]{function to sample $p$ percent elements\\in X according to the values in Y}\\
\bottomrule
\end{tabular}
\caption{Notations in Algorithm~\ref{alg:training}.}
\label{tab:notations}
\end{center}
\end{table}
\setlength{\tabcolsep}{1.4pt}

\begin{algorithm}[t]
\caption {DaR Training Policy. $\mathcal{W}$: number of warm-up epochs, $\mathcal{I}$: number of interval epochs, $p$: percentage of samples to keep, $\alpha$: number of active epochs, $\mathcal{R}$: list of epochs to reuse all data. Other notations are in Table~\ref{tab:notations}.}{\algblue{Blue part} is the standard training loop.\\\algred{Red part} is our method. (Best viewd in color)}
% {
% $\mathcal{X}$ is the set of all training images.
% $\mathcal{Y}$ is the set of all training labels.
% $E$ is the number of training epochs. $W$ is the number of warm-up epochs where all training data is used at the beginning of the training. $I$ is the subsampling interval. $p$ is the percentage of images to keep (\emph{w.r.t.} remaining training images). $\alpha$ is the number of active epochs where subsampling is performed. \textcolor{blue}{Blue part} is the standard training loop. \textcolor{red}{Red part} is our method.}
\label{alg:training}
\begin{algorithmic}
\State \textbf{input}: $\mathcal{X} = \{x_1, ..,x_N\} $, $\mathcal{Y} = \{y_1, ..,y_N\}$
\State \textbf{input}: $\mathcal{E}$, $\mathcal{W}$, $\mathcal{I}$, $p$, $\alpha$, $\mathcal{R}$
\State $\mathcal{T}_\text{start} = \mathcal{W}$, $\mathcal{T}_\text{previous} = \mathcal{W}$
\State$\mathcal{X}_{\text{s}} \leftarrow \mathcal{X}$
\State$\mathcal{Y}_{\text{s}} \leftarrow \mathcal{Y}$
\For {$\tau = 1, ..., \mathcal{E}$}
\State$\mathcal{L} \leftarrow \{\}$
\For {every batch $b$}
\State \algblue{$\mathcal{O}_b \leftarrow \text{forward}(\mathcal{X}_{\text{s},b})$}
\State \algblue{$\mathcal{L}_b \leftarrow \text{loss}(\mathcal{O}_b, \mathcal{Y}_{\text{s},b})$}
\State \algblue{$\text{backward}(\mathcal{L}_b)$}
\State \algblue{Update model weights}
\EndFor
% }
\algred{
\If{$\tau > \mathcal{W}$}
\If{$\tau - \mathcal{T}_\text{previous} = \mathcal{I}$ \textbf{and} $\tau - \mathcal{T}_\text{start} < \alpha$}
\State $\mathcal{T}_\text{previous} \leftarrow \tau$
\State $\mathcal{L} \leftarrow \text{sort}(\mathcal{L})$
\State $(\mathcal{X}_{\text{s}}, \mathcal{Y}_{\text{s}}) \leftarrow \text{sample}((\mathcal{X}_{\text{s}}, \mathcal{Y}_{\text{s}}), p, \mathcal{L})$
% \State $\mathcal{Y}_{\text{s}} \leftarrow \text{sample}(\mathcal{Y}_{\text{s}}, p, \mathcal{L})$
\EndIf
\EndIf
\If{$\tau \in \mathcal{R}$}
\State$\mathcal{T}_\text{start} \leftarrow \tau$
\State$\mathcal{X}_{\text{s}} \leftarrow \mathcal{X}$
\State$\mathcal{Y}_{\text{s}} \leftarrow \mathcal{Y}$
\EndIf
}
\EndFor
% \State\textbf{return} {$\mathcal{D}, \mathcal{S}$}
\end{algorithmic}
\end{algorithm}
\subsection{Motivation}
When human beings begin to learn and develop their intelligence, they (common people) tend to start with learning a broad area of knowledge without digging into the depth. For example, in the early stage of school, people take a variety of courses in different subjects (\emph{e.g.} math, physics and chemistry) at the introduction level. As the learning process proceeds, people will focus on a narrower and much more difficult area (\emph{e.g.} calculus in math). During this process, people start to forget what they have learned before because they do not use this knowledge often (\emph{e.g.} physics, chemistry). However, if people want to catch up materials that they have forgotten, they can do it in a much shorter time and usually, they can even learn better.  %\tiger{This repeats from the introduction, and I don't think this is needed especially for the method part. }
 
The process of this learning method is extremely efficient, as people gradually focus on less and less ``hard'' knowledge. This strategy (often) adopted by humans motivates this work. Specifically, we ask the question that whether deep neural networks can be trained efficiently by imitating the learning method of human beings. The goal of this work is to train a model with as less computation as possible while simultaneously maintaining the performance in classification and without decreasing the generalization ability of the learned parameters in the downstream vision tasks.
%Without loss of generality, we demonstrate that the goal is achievable in the task of image classification.

% \subsection{Abstraction of Human Learning Process}
We begin by abstracting the human learning process\footnote{Note that it \textbf{does not} necessarily mean human truly learns in this way from neuron-science perspective.}. The broad area of knowledge human learned during their early stage is similar to feeding all training data to the model at the start of the training iterations. As human begin to focus on a more difficult subject, we can imitate this process by gradually feeding only the most difficult examples (examples with large losses) to the model. Since human periodically review what they have learned before, we also periodically feed all the training data again to the model and restart the above process. A detailed training method will be discussed in the following.

\subsection{Drop-and-Refresh Training}
\label{sec:policy}
A high-level explanation of DaR is shown in Figure~\ref{fig:training_policy}.

%In this section, we first summarize the set of training parameters (listed in Table~\ref{tab:notations}) and their notations that will be used throughout this paper. Then, the details of how to conduct the training of ``Drop-and-Pick" (DaP) are provided.

\subsubsection{Notations} A list of notations is shown in Table~\ref{tab:notations}. We set a number of ``warm-up'' epochs ($\mathcal{W}$) at the beginning of training. In this ``warm-up'' period, the model is guaranteed to see all training examples. This is an imitation of human taking a broad range of courses. Then, we use a combination of number of interval epochs ($\mathcal{I}$) and a keep rate ($p$) to control the length of a learning stage (subsample examples every $\mathcal{I}$ epochs) and the number of remaining examples (keep $p\%$ of the current examples) respectively. Furthermore, a number of active epochs ($\alpha$) is used to control the number of learning stages we have (stop sampling after $\alpha$ epochs). Finally, we set a list of epochs $\mathcal{R}$ to reuse all training examples to prevent forgetting. %\tiger{My feeling is that introducing these parameters while explaining the training procedure will be better.}

\subsubsection{Training Details} Our DaR can be summarized as the process of periodically dropping easy examples and picking up all examples. As shown in Algorithm~\ref{alg:training}, we divide the overall training process into different cycles/periods denoted by $\mathcal{R}$. In each period, we keep $p\%$ of the hardest examples every $\mathcal{I}$ epochs (the ``drop'' stage). To find the $p\%$ of the hardest examples, we first sort examples by their losses in descending order and keep the top-$p\%$ examples. When the training process goes to a new period/cycle, we reuse all training examples (the ``refresh'' stage) followed by another ``drop'' stage. The only exception is the first period where we set a warm-up period $\mathcal{W}$ to use all examples to ``warm-up'' the model. During the first $\mathcal{W}$ epochs, we keep all examples for training. We also set an optional ``active'' epochs $\alpha$ because we do not want to drop too much (\eg $90\%$) training data when the period is long. If an $\alpha$ is set, then the ``drop'' stage only lasts for $\alpha$ epochs within each training cycle.
% Our DaP can be summarized (in Algorithm~\ref{alg:training}) as the process of periodically dropping easy examples and picking up all examples. Specificially, we divide the overall training process into different cycles/periods and save the beginning epoch of each cycles/periods into a list $\mathcal{R}$. In each period, we keep $p\%$ of the hardest examples every $\mathcal{I}$ epochs (the ``drop'' stage). To find the $p\%$ of the hardest examples, we first sort examples by their losses from last epoch in descending order and keep the top-$p\%$ examples. When the training process goes to a new period/cycle, we reuse all training examples (the ``pick'' stage) followed by another ``drop'' stage. The only exception is the first period where we set a warm-up period $\mathcal{W}$ to use all examples to ``warm-up'' the model. During the first $\mathcal{W}$ epochs, we keep all examples for training. We also set an optional ``active'' epochs $\alpha$. If an $\alpha$ is set, then the ``drop'' stage only lasts for $\alpha$ epochs within each training cycle. And setting $\alpha$ to $\infty$ means we keep dropping samples in each period.

In our experiments, we find that changing these hyperparameters in a reasonable range does not make much difference. Thus we simply set hyperparameters related to DaR for ImageNet classification as follow. Since models are usually trained for $90 \sim 120$ epochs, we set warmup epochs to around $10\%$ if the total training epochs which is $10$ epochs. More over, we set interval=2, percentage=0.9, active=10, reusing all data at $30^{th}, 60^{th}, 90^{th}$ epochs which corresponds to reusing all data when learning rate is dropped. All other hyperparameters for training (\eg optimizer, learning rate, weight decay) are kept same as models' original ones.

\section{Main Results}

In this section, we present main results on multiple common image classification benchmarks with state-of-the-art architectures to demonstrate that our proposed DaR method using non-\iid data has the potential to achieve better performance than uniform sampling using \iid data. Moreover, we also show that DaR pre-trained models generalize well to downstream tasks.
\begin{table}[h]\setlength{\tabcolsep}{1pt}
	\centering
    \resizebox{0.4\textwidth}{!}{
    \begin{tabular}{l|c|c|c}
     CIFAR10& ori & re-imp & 83\% cost \\
    \hline
    ResNet-110 \cite{he2016deep} & 93.57 & 94.54 & $95.08_{(+0.54)}$ \\
    ResNet-164 \cite{he2016deep} & -     & 95.47 & $95.61_{(+0.14)}$ \\
	\hline
	DenseNet-BC-100 \cite{huang2017densely} & 95.49 & 95.16 & $95.43_{(+0.27)}$ \\
% 	\hline\hline
% 	 CIFAR100 & ori & re-imp & 92\% comp \\
%     \hline
%     ResNet-110 \cite{he2016deep} & -     & 74.26 & $74.43_{(+0.17)}$ \\
%     ResNet-164 \cite{he2016deep} & 74.84 & 77.29 & $76.84_{(-0.45)}$ \\
% 	\hline
% 	DenseNet-BC-100 \cite{huang2017densely} & 77.73 & 76.88 & $77.20_{(+0.32)}$ \\
% % 	ResNext-29-8x64d \cite{xie2017aggregated} & 96.35 & 96.33 & 96.22 & 96.29 \\
		\end{tabular}
        }
		\caption{Comparisons on CIFAR10 ($\%$).}
        % Comparison with state-of-the-arts reported in recent publications with different backbones.
\label{tab:cifar10_networks}
%  \vspace{-4mm}
\end{table}

% \begin{table}[t!]\setlength{\tabcolsep}{1pt}
% 	\centering
%     \resizebox{0.49\textwidth}{!}{
%     \begin{tabular}{l|c|c|c|c}
%      CIFAR10& ori & re-imp & 63\% comp & 83\% comp \\
%     \hline
%     ResNet-110 \cite{he2016deep} & 93.57 & 94.54 & $94.63_{(+0.09)}$ & $95.08_{(+0.54)}$ \\
%     ResNet-164 \cite{he2016deep} & -     & 95.47 & $95.21_{(-0.26)}$ & $95.61_{(+0.14)}$ \\
% 	\hline
% 	DenseNet-BC-100 \cite{huang2017densely} & 95.49 & 95.16 & $94.65_{(-0.51)}$ & $95.43_{(+0.27)}$ \\
% 	\hline\hline
% 	 CIFAR100 & ori & re-imp & 79\% comp & 92\% comp \\
%     \hline
%     ResNet-110 \cite{he2016deep} & -     & 74.26 & $74.05_{(-0.21)}$ & $74.43_{(+0.17)}$ \\
%     ResNet-164 \cite{he2016deep} & 74.84 & 77.29 & $76.51_{(-0.78)}$ & $76.84_{(-0.45)}$ \\
% 	\hline
% 	DenseNet-BC-100 \cite{huang2017densely} & 77.73 & 76.88 & $77.03_{(+0.15)}$ & $77.20_{(+0.32)}$ \\
% % 	ResNext-29-8x64d \cite{xie2017aggregated} & 96.35 & 96.33 & 96.22 & 96.29 \\
% 		\end{tabular}
%         }
% 		\caption{Comparisons on CIFAR10 and CIFAR100 ($\%$).}
%         % Comparison with state-of-the-arts reported in recent publications with different backbones.
% \label{tab:cifar10_networks}
%  \vspace{-4mm}
% \end{table}

\begin{table}[h]\setlength{\tabcolsep}{1pt}
	\centering
    \resizebox{0.4\textwidth}{!}{
    \begin{tabular}{l|c|c|c}
    %  CIFAR10& ori & re-imp & 83\% cost \\
    % \hline
%     ResNet-110 \cite{he2016deep} & 93.57 & 94.54 & $95.08_{(+0.54)}$ \\
%     ResNet-164 \cite{he2016deep} & -     & 95.47 & $95.61_{(+0.14)}$ \\
% 	\hline
% 	DenseNet-BC-100 \cite{huang2017densely} & 95.49 & 95.16 & $95.43_{(+0.27)}$ \\
% 	\hline\hline
	 CIFAR100 & ori & re-imp & 92\% cost \\
    \hline
    ResNet-110 \cite{he2016deep} & -     & 74.26 & $74.43_{(+0.17)}$ \\
    ResNet-164 \cite{he2016deep} & 74.84 & 77.29 & $76.84_{(-0.45)}$ \\
	\hline
	DenseNet-BC-100 \cite{huang2017densely} & 77.73 & 76.88 & $77.20_{(+0.32)}$ \\
% 	ResNext-29-8x64d \cite{xie2017aggregated} & 96.35 & 96.33 & 96.22 & 96.29 \\
		\end{tabular}
        }
		\caption{Comparisons on CIFAR100 ($\%$).}
        % Comparison with state-of-the-arts reported in recent publications with different backbones.
\label{tab:cifar100_networks}
%  \vspace{-4mm}
\end{table}

% \begin{table}[htb]\setlength{\tabcolsep}{1pt}
% 	\centering
%     \resizebox{0.49\textwidth}{!}{
%     \begin{tabular}{l|c|c|c|c}
%      & original & re-implement & 79\% comp & 92\% comp \\
%     \hline
%     ResNet-110 \cite{he2016deep} & -     & 74.26 & $74.05_{(-0.21)}$ & $74.43_{(+0.17)}$ \\
%     ResNet-164 \cite{he2016deep} & 74.84 & 77.29 & $76.51_{(-0.78)}$ & $76.84_{(-0.45)}$ \\
% 	\hline
% 	DenseNet-BC-100 \cite{huang2017densely} & 77.73 & 76.88 & $77.03_{(+0.15)}$ & $77.20_{(+0.32)}$ \\
% 	\hline
% % 	ResNext-29-8x64d \cite{xie2017aggregated} & 82.23 & 82.78 & 81.94 & 82.79 \\
% 		\end{tabular}
%         }
% 		\caption{CIFAR100 accuracy ($\%$).}
%         % Comparison with state-of-the-arts reported in recent publications with different backbones.
% \label{tab:cifar100_networks}
%  \vspace{-4mm}
% \end{table}
\begin{table*}[t]
\renewcommand\arraystretch{1.1}
\captionsetup{font=small}
\newcommand{\tabincell}[2]{\begin{tabular}{@{}#1@{}}#2\end{tabular}}
\begin{center} {\scalebox{0.95}{
\begin{tabular}{l|p{1.8cm}<{\centering}|p{1.8cm}<{\centering}|p{1.8cm}<{\centering}|p{1.8cm}<{\centering}|p{2.4cm}<{\centering}|p{2.4cm}<{\centering}}
\hline
\multirow{2}{*}{} & \multicolumn{2}{c|}{original} & \multicolumn{2}{c|}{re-implementation} & \multicolumn{2}{c}{86\% cost}\\
\cline{2-7}
& \tabincell{c}{top-1 err.} & \tabincell{c}{top-5 err.} & \tabincell{c}{top-1err.} & \tabincell{c}{top-5 err.} & \tabincell{c}{top-1 err.} & \tabincell{c}{top-5 err.} \\
\hline
ResNet-18 \cite{he2016deep} & - & - & $30.45$ & $10.95$ & $29.90_{(\textcolor{red}{+0.55})}$  & $10.70_{(\textcolor{red}{+0.25})}$ \\
ResNet-50 \cite{he2016deep} & $24.7$ & $7.8$ & $24.47$ & $7.25$ & $24.38_{(\textcolor{red}{+0.09})}$  & $7.27_{(\textcolor{blue}{-0.02})}$ \\
ResNet-101 \cite{he2016deep} & $23.6$ & $7.1$ & $23.12$ & $6.51$ & $22.87_{(\textcolor{red}{+0.25})}$ & $6.44_{(\textcolor{red}{+0.07})}$\\
ResNet-152 \cite{he2016deep} & $23.0$ & $6.7$ & $22.42$ & $6.34$ & $22.44_{(\textcolor{blue}{-0.02})}$ & $6.25_{(\textcolor{red}{+0.09})}$\\
% \hline
% ResNeXt-50 \cite{xie2017aggregated} & $22.2$ & - & $22.11$ & $5.90$ & - & - & - & -\\
% ResNeXt-101 \cite{xie2017aggregated} & $21.2$ & $5.6$ & $21.18$ & $5.57$ & - & - & - & -\\
\hline
DenseNet-121 \cite{huang2017densely} & $25.02$ & $7.71$ & $24.69$ & $7.55$ & $24.51_{(\textcolor{red}{+0.18})}$ & $7.37_{(\textcolor{red}{+0.18})}$ \\
DenseNet-169 \cite{huang2017densely} & $23.80$ & $6.85$ & $23.34$ & $6.68$ & $23.39_{(\textcolor{blue}{-0.05})}$ & $6.69_{(\textcolor{blue}{-0.01})}$\\
% DenseNet-201 \cite{huang2017densely} & $22.58$ & $6.34$ & - & - & - & - & -  & -\\
% DenseNet-161 \cite{huang2017densely} & $22.33$ & $6.15$ & - & - & - & - & -  & -\\
\hline
MobileNet-V1-1.0 \cite{howard2017mobilenets} & $29.4$ & - & $28.91$ & $10.15$ & $28.59_{(\textcolor{red}{+0.32})}$ & $10.16_{(\textcolor{blue}{-0.01})}$\\
% MobileNet-V1-0.75 \cite{howard2017mobilenets} & $31.6$ & - & $30.92$ & $11.44$ & $31.84_{(-0.92)}$ & $11.48_{(-0.04)}$ & $31.11_{(-0.19)}$ & $11.44_{(0.00)}$\\
% MobileNet-V1-0.5 \cite{howard2017mobilenets} & $36.3$ & - & - & - & - & - & - & -\\
% MobileNet-V1-0.25 \cite{howard2017mobilenets} & $49.4$ & - & - & - & - & - & - & -\\
\hline
\end{tabular}}}
\end{center}
% \vspace{-3mm}
\caption{\label{tab:imagenet-results}Single-crop error rates (\%) on the ImageNet validation set and training computation comparisons. The \textit{original} column refers to the results reported in the original papers. To enable a fair comparison, we re-train the baseline models and report the scores in the \textit{re-implementation} column. The last two columns refers to different training methods with different computation compared with baselines. The numbers in brackets denote the performance improvement over the re-implemented baselines.}
\end{table*}
\begin{table*}[tb]\setlength{\tabcolsep}{1pt}
\captionsetup{font=small}
	\centering
% 	\tiny
    % \resizebox{1.0\textwidth}{!}{
\begin{tabular}{l|l|c|ccc|ccc|ccc|ccc}
 Method & Backbone & Pre-train Cost & $\text{AP}^{\text{bb}}$ & $\text{AP}_{50}^{\text{bb}}$ & $\text{AP}_{75}^{\text{bb}}$ & $\text{AP}_{\text{S}}^{\text{bb}}$ & $\text{AP}_{\text{M}}^{\text{bb}}$ & $\text{AP}_{\text{L}}^{\text{bb}}$ & $\text{AP}^{\text{m}}$ & $\text{AP}_{50}^{\text{m}}$ & $\text{AP}_{75}^{\text{m}}$ & $\text{AP}_{\text{S}}^{\text{m}}$ & $\text{AP}_{\text{M}}^{\text{m}}$ & $\text{AP}_{\text{L}}^{\text{m}}$ \\
        \hline
        
     FPN \cite{lin2017feature} & ResNet-50 & 100\% & 35.8 & 57.7 & 38.2 & 20.2 & 39.5 & 45.9 & - & - & - & - & - & - \\
    %  FPN \cite{lin2017feature} & ResNet-50 & 75\% & \textcolor{blue}{35.9} & \textcolor{blue}{57.8} & \textcolor{blue}{38.4} & \textcolor{red}{21.1} & \textcolor{blue}{39.7} & \textcolor{blue}{46.6} & - & - & - & - & - & -\\
    %  FPN \cite{lin2017feature} & ResNet-50 & 86\% & \textcolor{red}{36.1} & \textcolor{red}{57.9} & \textcolor{red}{38.9} & \textcolor{blue}{20.6} & \textcolor{red}{39.9} & \textcolor{red}{46.9} & - & - & - & - & - & - \\
    FPN \cite{lin2017feature} & ResNet-50 & 86\% & \textbf{36.1} & \textbf{57.9} & \textbf{38.9} & \textbf{20.6} & \textbf{39.9} & \textbf{46.9} & - & - & - & - & - & - \\
     $\Delta$ &  & $-14\%$ & $+0.3$ & $+0.2$ & $+0.7$ & $+0.4$ & $+0.4$ & $+1.0$ & - & - & - & - & - & - \\
    %  FPN \cite{lin2017feature} & ResNet-101 & 100\% & 38.5 & 59.9 & 42.0 & 22.2 & 42.4 & 50.0 & - & - & - & - & - & - \\
    % \hline
    %  FPN \cite{lin2017feature} & ResNet-101 & 100\% & 38.4 & 60.1 & 41.7 & 22.3 & 42.6 & 49.4 & - & - & - & - & - & - \\
    %  FPN \cite{lin2017feature} & ResNet-101 & 75\% & 38.4 & 60.3 & 41.6 & 22.6 & 42.6 & 49.5 & - & - & - & - & - & - \\
    %  FPN \cite{lin2017feature} & ResNet-101 & 86\% & 38.5 & 60.3 & 41.9 & 22.8 & 42.6 & 49.8 & - & - & - & - & - & - \\
     
        \hline
    
     Mask R-CNN \cite{he2017mask} & ResNet-50 & 100\% & 36.6 & 58.0 & 39.5 & 20.8 & 40.1 & 47.6 & 33.5 & 54.8 & 35.4 & 17.0 & 36.8 & 46.0 \\
    %  Mask R-CNN \cite{he2017mask} & ResNet-50 & 75\% & \textcolor{blue}{36.8} & \textcolor{blue}{58.4} & \textcolor{blue}{39.7} & \textcolor{red}{21.4} & \textcolor{red}{40.5} & \textcolor{blue}{47.7} & \textcolor{red}{33.8} & \textcolor{red}{55.3} & \textcolor{blue}{35.8} & \textcolor{red}{17.7} & \textcolor{red}{37.2} & \textcolor{red}{46.3}\\
    %  Mask R-CNN \cite{he2017mask} & ResNet-50 & 86\% & \textcolor{red}{36.9} & \textcolor{red}{58.5} & \textcolor{red}{39.9} & \textcolor{blue}{21.0} & \textcolor{red}{40.5} & \textcolor{red}{47.8} & \textcolor{red}{33.8} & \textcolor{blue}{55.1} & \textcolor{red}{35.9} & \textcolor{blue}{17.5} & \textcolor{blue}{37.1} & \textcolor{blue}{46.2} \\
    Mask R-CNN \cite{he2017mask} & ResNet-50 & 86\% & \textbf{36.9} & \textbf{58.5} & \textbf{39.9} & \textbf{21.0} & \textbf{40.5} & \textbf{47.8} & \textbf{33.8} & \textbf{55.1} & \textbf{35.9} & \textbf{17.5} & \textbf{37.1} & \textbf{46.2} \\
     $\Delta$ &  & $-14\%$ & $+0.3$ & $+0.5$ & $+0.4$ & $+0.3$ & $+0.4$ & $+0.2$ & $+0.3$ & $+0.3$ & $+0.5$ & $+0.5$ & $+0.3$ & $+0.2$ \\
    %  \hline
    %  Mask R-CNN \cite{he2017mask} & ResNet-101 & 100\% & 39.5 & 60.9 & 43.0 & 22.8 & 43.7 & 51.0 & 35.8 & 57.5 & 38.2 & 18.8 & 39.5 & 48.5 \\
    %  Mask R-CNN \cite{he2017mask} & ResNet-101 & 75\% & - & - & - & - & - & - & - & - & - & - & - & - \\
    %  Mask R-CNN \cite{he2017mask} & ResNet-101 & 86\% & - & - & - & - & - & - & - & - & - & - & - & - \\
	
		\end{tabular}
        % }
% 		\caption{\textbf{COCO2017 \texttt{val} detection and segmentation results.} $\text{AP}^{\text{bb}}$ denotes box AP, $\text{AP}^{\text{m}}$ denotes mask AP. \textcolor{red}{Red}: best, \textcolor{blue}{Blue}: second best.
% 		In particular, our results using $75\%$ training computation consistently outperform the ones using $86\%$ and $100\%$ on small objects, which suggest \textbf{our models have successfully transferred the knowledge learned on hard cases during pre-training}.
% 		}
        \caption{\textbf{COCO2017 \texttt{val} detection and segmentation results.} $\text{AP}^{\text{bb}}$ denotes box AP, $\text{AP}^{\text{m}}$ denotes mask AP. 
		In particular, our results using $86\%$ training computation consistently outperform the ones using $100\%$ on all metric, which suggest \textbf{our models have successfully transferred the knowledge learned on hard cases during pre-training}.
		}
% 		\vspace{-5mm}
        % All detectors use ResNet-101 as backbone and DCR modules use ResNet-152 as base classifier.
\label{tab:coco_detection}
% \vspace{-2mm}
\end{table*}

\subsection{Implementation Details}
All models in this paper are implemented using PyTorch \cite{paszke2017automatic}. If otherwise stated, models on the CIFAR datasets are trained with 300 epochs, learning rate is divided by 10 at 150 and 225 epoch respectively. On the ImageNet dataset, models are trained with 120 epochs, dividing the learning rate by 10 every 30 epochs. We follow the batchsize, learning rate, weight decay settings in their original implementations.

During the training, we accumulate the loss of each image on the fly \textbf{with data augmentation} at each training iteration. When we sample a subset of training example, we simply sort examples according to their losses in descending order and keep the top $p\%$ examples to form our new training set. Thus, this sampling method is non-\iid. When we report the training computation, we set the strategy of using all data as comparison baseline (100\% computation). And we compute the ratio of between total examples used by our proposed method with the baseline as the computation for our method.

\subsection{Non-\iid Sampling in Image Classification}
Image classification is the most fundamental task in computer vision and it receives most attention. However, most of the recent efforts mostly focus on either network designs~\cite{szegedy2015going,he2016deep,xie2017aggregated,huang2017densely,sandler2018mobilenetv2,ma2018shufflenetv2} or regularization~\cite{szegedy2016rethinking,zhang2017mixup} during training and uniformly sample \iid training data for granted. In this section, we empirically show that our DaR sampling method with non-\iid training data has the potential of achieving better performance.

\subsubsection{CIFAR}
CIFAR10 and CIFAR100 datasets \cite{krizhevsky2009cifardataset} both consist of colored natural images with $32\times32$ pixels. CIFAR10 consists of images with 10 classes and CIFAR100 has 100 classes. The training and test sets contain 50,000 and 10,000 images respectively for both datasets. We adopt standard data augmentations (random crop and random flip) and normalize the data with the channel means and standard deviations. We train our models with 50,000 training images and report the test accuracy on the test set.

To validate our proposed method can be generalized to other architectures instead of ResNet-110, we perform experiments on three more state-of-the-art architecutures: ResNet-164 \cite{he2016deep} and DenseNet-BC-100 \cite{huang2017densely} on both CIFAR10 and CIFAR100 datasets with 83\% and 92\% cost respectively. Results are shown in Table~\ref{tab:cifar10_networks} and Tabel~\ref{tab:cifar100_networks}. On all these architectures, our method achieves on-par or even better results on both datasets. 

% \begin{table}[tb]\setlength{\tabcolsep}{1pt}
% 	\centering
% % 	\tiny
%     % \resizebox{1.0\textwidth}{!}{
% \begin{tabular}{l|l|c|c}
%  Method & Backbone & Pre-train Computation & AP  \\
%         \hline
        
%      SimpleBaseline \cite{xiao2018simple} & ResNet-50 & 100\% & 70.4 \\
%      SimpleBaseline \cite{xiao2018simple} & ResNet-50 & 74.9\% & 70.7 \\
%      SimpleBaseline \cite{xiao2018simple} & ResNet-50 & 86.0\% & 70.5 \\
     
%         \hline
    
%      SimpleBaseline \cite{xiao2018simple} & ResNet-101 & 100\% & 72.0 \\
%      SimpleBaseline \cite{xiao2018simple} & ResNet-101 & 74.9\% & 72.4 \\
%      SimpleBaseline \cite{xiao2018simple} & ResNet-101 & 86.0\% & - \\
	
% 		\end{tabular}
%         % }
% 		\caption{\textbf{COCO2017 \texttt{val} keypoint results.}}
%         % All detectors use ResNet-101 as backbone and DCR modules use ResNet-152 as base classifier.
%     \label{tab:coco_keypoint}
% % \vspace{-2mm}
% \end{table}

\begin{table*}[!t]\setlength{\tabcolsep}{2pt}
\captionsetup{font=small}
	\centering
% 	\tiny
    % \resizebox{1.0\textwidth}{!}{
    \begin{center} {\scalebox{0.9}{
\begin{tabular}{l|l|c|ccc|cc|ccc|cc}
 Method & Backbone & Pre-train Cost & $\text{AP}^{\text{kp}}$ & $\text{AP}_{50}^{\text{kp}}$  & $\text{AP}_{75}^{\text{kp}}$ & $\text{AP}_{\text{M}}^{\text{kp}}$  & $\text{AP}_{\text{L}}^{\text{kp}}$ & $\text{AR}^{\text{kp}}$ & $\text{AR}_{50}^{\text{kp}}$  & $\text{AR}_{75}^{\text{kp}}$ & $\text{AR}_{\text{M}}^{\text{kp}}$  & $\text{AR}_{\text{L}}^{\text{kp}}$\\
        \hline
        
    %  SimpleBaseline \cite{xiao2018simple} & ResNet-50 & 100\% & 70.4 & 91.4 & \textcolor{red}{78.2} & \textcolor{blue}{67.7} & 74.4 & 73.5 & 92.1 & \textcolor{blue}{80.5} & \textcolor{blue}{70.4} & 78.3\\
    %  SimpleBaseline \cite{xiao2018simple} & ResNet-50 & 75\% & \textcolor{red}{70.7} & \textcolor{red}{91.5} & \textcolor{red}{78.2} & \textcolor{red}{67.9} & \textcolor{red}{75.1} & \textcolor{red}{74.0} & \textcolor{red}{92.6} & \textcolor{red}{80.7} & \textcolor{red}{70.7} & \textcolor{red}{78.9}\\
    %  SimpleBaseline \cite{xiao2018simple} & ResNet-50 & 86\% & \textcolor{blue}{70.5} & \textcolor{red}{91.5} & 78.1 & 67.5 & \textcolor{blue}{74.9} & \textcolor{blue}{73.7} & \textcolor{blue}{92.3} & 80.3 & 70.3 & \textcolor{blue}{78.7}\\
    
    % SimpleBaseline \cite{xiao2018simple} & ResNet-50 & 100\% & 70.4 & 91.4 & \textbf{78.2} & \textbf{67.7} & 74.4 & 73.5 & 92.1 & \textbf{80.5} & \textbf{70.4} & 78.3\\
    %  SimpleBaseline \cite{xiao2018simple} & ResNet-50 & 86\% & \textbf{70.5} & \textbf{91.5} & 78.1 & 67.5 & \textbf{74.9} & \textbf{73.7} & \textbf{92.3} & 80.3 & 70.3 & \textbf{78.7}\\
     
    %     \hline
    
     SimpleBaseline \cite{xiao2018simple} & ResNet-101 & 100\% & 72.0 & 91.5 & 79.4 & 69.2 & 76.4 & 75.3 & 93.0 & 82.0 & 72.0 & 80.2\\
    %  SimpleBaseline \cite{xiao2018simple} & ResNet-101 & 75\% & \textcolor{blue}{72.4} & 91.5 & \textcolor{red}{80.3} & \textcolor{blue}{69.4} & \textcolor{blue}{76.8} & \textcolor{blue}{75.6} & \textcolor{blue}{92.8} & \textcolor{red}{82.4} & \textcolor{red}{72.4} & \textcolor{blue}{80.5}\\
    %  SimpleBaseline \cite{xiao2018simple} & ResNet-101 & 86\% & \textcolor{red}{72.6} & \textcolor{red}{92.5} & \textcolor{red}{80.3} & \textcolor{red}{69.6} & \textcolor{red}{77.0} & \textcolor{red}{75.7} & \textcolor{red}{93.1} & \textcolor{red}{82.4} & \textcolor{red}{72.4} & \textcolor{red}{80.7}\\
    SimpleBaseline \cite{xiao2018simple} & ResNet-101 & 86\% & \textbf{72.6} & \textbf{92.5} & \textbf{80.3} & \textbf{69.6} & \textbf{77.0} & \textbf{75.7} & \textbf{93.1} & \textbf{82.4} & \textbf{72.4} & \textbf{80.7}\\
    $\Delta$ &  & $-14\%$ & $+0.6$ & $+1.0$ & $+0.9$ & $+0.4$ & $+0.6$ & $+0.4$ & $+0.1$ & $+0.4$ & $+0.4$ & $+0.5$\\
	
		\end{tabular}}}
\end{center}
        % }
% 		\caption{\textbf{COCO2017 \texttt{val} keypoint results.} $\text{AP}^{\text{kp}}$ denotes keypoint AP, $\text{AR}^{\text{kp}}$ denotes keypoint AR.
% 		\textcolor{red}{Red}: best, \textcolor{blue}{Blue}: second best.
% 		The keypoint results are generally in accordance with those in Table~\ref{tab:coco_detection} in terms of the \textbf{knowledge transferability} of our pre-trained models.
% 		}
        \caption{\textbf{COCO2017 \texttt{val} keypoint results.} $\text{AP}^{\text{kp}}$ denotes keypoint AP, $\text{AR}^{\text{kp}}$ denotes keypoint AR.
		The keypoint results are generally in accordance with those in Table~\ref{tab:coco_detection} in terms of the \textbf{knowledge transferability} of our pre-trained models.
		}
% 		\vspace{-5mm}
        % All detectors use ResNet-101 as backbone and DCR modules use ResNet-152 as base classifier.
    \label{tab:coco_keypoint}
% \vspace{-2mm}
\end{table*}

\subsubsection{ImageNet}
The ImageNet2012 classification dataset \cite{deng2009imagenet} consists 1,281,167 images for training and 50,000 for validation with 1,000 classes. Following \cite{szegedy2015going}, we adopt random crop augmentation with patches covering $8\% \sim 100\%$ of the entire image, aspect ratio augmentation of [$\frac{3}{4}$, $\frac{4}{3}$], and random horizontal flip augmentation. All patches are resized to $224\times224$ and normalize the data with the channel means and standard deviations. We report single-crop classification errors on the validation set for fair comparison.

We apply our method with state-of-the-art architectures: ResNet \cite{he2016deep}, DenseNet \cite{huang2017densely}, and MobileNet \cite{howard2017mobilenets}. Results are shown in Table~\ref{tab:imagenet-results}. We report both our re-implementation results and original results in their papers and our reproduced results are always better than original ones. For a fair comparison, we only compare our method with the reproduced results.
When we use the set of default hyper-parameters we find for ImageNet, the cost of the training is only 86\% of that when all data is used. We observe that models trained with DaR sampling outperforms the uniform sampling counterpart in most of the cases which supports our hypothesis that sampling non-\iid training data might have the potential of achieving better performance.

\subsection{Generalization to Downstream Tasks}

Over the years, ImageNet2012 \cite{deng2009imagenet} has become the standard dataset to pre-train models for multiple downstream tasks (\emph{e.g.} object detection \cite{cheng2018revisiting, wei2018ts2c,cheng18decoupled,zhang2016faster,li2018scale}, segmentation \cite{zhao2017pyramid,deeplabv3plus2018,liang2016semantic,zhang2018context,lin2017refinenet}, human pose estimation \cite{xiao2018simple,sun2017integral,newell2016stacked,cao2017realtime,wei2016convolutional}). To demonstrate that our approach can generalize well to these downstream tasks and can even transfer knowledge distilled during pre-training, we evaluate our pre-trained models a wide range of downstream tasks including object detection, instance segmentation, scene parsing and human pose estimation. 

\subsubsection{Object Detection and Instance Segmentation}
We evaluate the performance for object detection and instance segmentation on the COCO2017 dataset \cite{lin2014microsoft}. COCO2017 has 115k training images, 5k validation images and 20k test image. We train FPN \cite{lin2017feature} (for object detection) and Mask R-CNN \cite{he2017mask} (for instance segmentation) on the 115k training set and evaluate the performance on the 5k validation set. We use the open-sourced framework mmdetection \cite{mmdetection2018}. We use a single scale of [800, 1333] during training and testing.

Results are shown in Table~\ref{tab:coco_detection}. 
% For a fair comparison, all models are initialized with PyTorch pre-trained models respectively. We notice that the baseline (pre-train cost: 100\%) result is lower than the one reported in \cite{lin2017feature,he2017mask} (35.8 compared with 36.7). This is because \cite{lin2017feature,he2017mask} use the original Caffe models released by \cite{he2016deep} whose mean and variance of BatchNorm layers are strictly computed using average (not moving average) on a sufficiently large training batch after the training procedure. Since we fix BatchNorm layer during training, the re-calibrated mean and variance is important to produce better result. We do not re-calibrate BatchNorm statistics as we do not know how \cite{he2016deep} exactly evaluate mean and variance.
For both object detection task and instance segmentation task, models pre-trained with our method (86\% cost) perform slightly better than models pre-trained with all data (100\% cost) in terms of the AP metric. This means knowledge learned in our method can be transfered to other tasks.

\subsubsection{Human Pose Estimation}
We further experiment the performance on human pose estimation task using the COCO2017 keypoint dataset \cite{lin2014microsoft}. 
It contains more than 200k images and 250k person instances labeled with keypoints. 150k instances of them are publicly available for training and validation. We only use COCO2017 train (57K images and 150K person instances) and evaluate the performance on the COCO2017 val set. We use the same implementation in \cite{xiao2018simple} and use a single input size of $256 \times 192$ during training and testing.

Results are shown Table~\ref{tab:coco_keypoint}. All reported baseline results are similar or even better than the ones reported in \cite{xiao2018simple}. We observe a consistent improvement in the human pose estimation task.
% Since the human pose estimation task do not consider small (less than $32\times32$ pixels) human instances, the hard case in this task is easier than that in object detection and semantic segmentation.
We further perform experiments with ResNet-101 backbone in human pose estimation. Interestingly, with 86\% pre-train cost, the performance of our DaR pre-trained model exceeds baseline by a large margin of 0.6 $\text{AP}^{\text{kp}}$. It shows that DaR pre-trained models have the potential to further improve human pose estimation.

\begin{table}[th]\setlength{\tabcolsep}{1pt}
	\centering
	%\tiny
	\small
    \resizebox{0.4\textwidth}{!}{
\begin{tabular}{l|l|c|c}
 Method & Backbone & \makecell[l]{Pre-train Cost} & mIoU  \\
        \hline

      PSPNet \cite{zhao2017pyramid} & ResNet-50 & 100\% & 76.5  \\
    %   PSPNet \cite{zhao2017pyramid} & ResNet-50 & 75\% & 76.8  \\
      PSPNet \cite{zhao2017pyramid} & ResNet-50 & 86\% & \textbf{77.0}  \\
      $\Delta$ &  & $-14\%$ & $+0.5$  \\
    %   \hline
      
    %   \hline
    %   PSPNet \cite{zhao2017pyramid} & ResNet-101 & 100\% & - & - & - & - & - & - & - & - & - & - & - & - & - & - & - & - & - & - & - & - \\
    %   PSPNet \cite{zhao2017pyramid} & ResNet-101 & 75\% & - & - & - & - & - & - & - & - & - & - & - & - & - & - & - & - & - & - & - & - \\
    %   PSPNet \cite{zhao2017pyramid} & ResNet-101 & 86\% & - & - & - & - & - & - & - & - & - & - & - & - & - & - & - & - & - & - & - & - \\
	
		\end{tabular}
        }
		\caption{\textbf{Cityscapes \texttt{val} scene parsing results.}}
\label{tab:cityscapes}
%  \vspace{-4mm}
\end{table}
\subsubsection{Scene Parsing}
We perform experiments on the Cityscapes dataset \cite{cordts2016cityscapes} using the PSPNet~\cite{zhao2017pyramid}. 
It is a large-scale dataset containing high quality pixel-level annotations (fine data) of 5000 images (2975, 500, and 1525 for the training, validation, and test sets respectively) and about 20000 coarsely annotated images (coarse data).
We train PSPNet \cite{zhao2017pyramid} with a publicly available toolbox \cite{torch2018segment} and only using the fine data and evaluate on the validation set. During training, we use a single scale of $769 \times 769$. Results are shown in Table~\ref{tab:cityscapes}. We find DaR pre-trained models also generalize well in the semantic segmentation task.

\begin{figure}%
\centering
\subfigure[Uniform sample.]{%
% \label{fig:first}%
\includegraphics[width=0.45\linewidth]{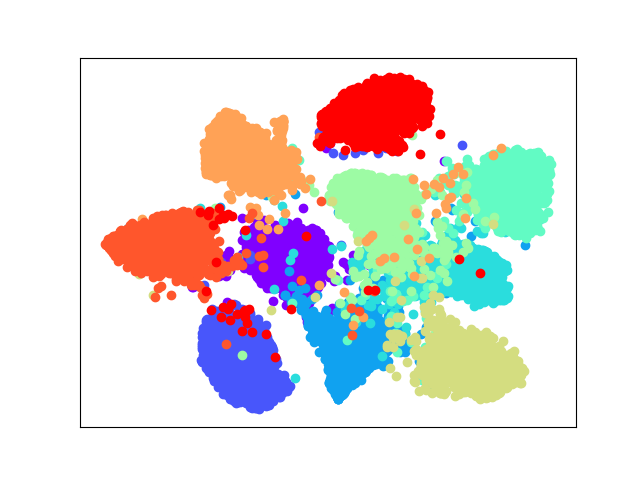}}%
% \qquad
~
\subfigure[DaR sample.]{%
% \label{fig:second}%
\includegraphics[width=0.45\linewidth]{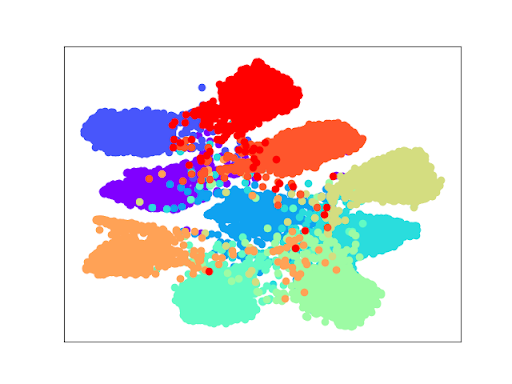}}%
\caption{
t-SNE visualization of ResNet-110 on CIFAR-10 dataset, each color represents a class. (a) Using uniform sample, training data is \iid. (b) Using DaR sample, training data is non-\iid.
}
\label{fig:t-sne}
\end{figure}

% \begin{figure}[t]
% 	\centering
% 	\begin{minipage}[b]{0.99\linewidth}
%     \centering
%     \includegraphics[width=0.86\linewidth]{figures/tSNE-resnet_v1_110-cifar10-all-data.png}
%     \\
%     {\footnotesize(a) Uniform sample.}
%     \\
%     \end{minipage}
%     \begin{minipage}[b]{0.99\linewidth}
%     \centering
%     \vspace{0mm}
%     \includegraphics[width=0.86\linewidth]{figures/DaR.png}
%     \\
%     {\footnotesize(b) DaR sample.}
%     \\
%     \end{minipage}
% 	\caption{
% 	t-SNE visualization of ResNet-110 on CIFAR-10 dataset, each color represents a class. (a) Using uniform sample, training data is \iid. (b) Using DaR sample, training data is non-\iid.
% 	}
% % 	\vspace{-5mm}
% 	\label{fig:t-sne}
% \end{figure}

\section{Discussions}

In summary, our experiments demonstrate that the DaR sampling strategy can achieve comparable (in many cases better) predictive accuracy with less training cost on various datasets (CIFAR 10, CIFAR 100 and ImageNet) and with different backbone architectures (ResNets, DenseNets and MobileNets). More importantly, we find that the ImageNet pre-trained models using our DaR sampling strategy can achieve much better transferability for the downstream tasks including object detection, instance segmentation, scene parsing and human pose estimation consistently. Based on these observations, in this section we try to initiate some discussions which hopefully can motivate people to rethink the connections between the sampling strategy for training and the transferability of its learned features for pre-training ImageNet models.

% \textbf{\emph{Should we stop the \iid sampling strategy when training image classification networks?}} Although the final sampled data is non-\iid, we cannot conclude the training dataset should not necessarily be \iid. This is because our non-\iid data is still sampled from a set of \iid data.

% \textbf{\emph{Is ImageNet pre-training still necessary as a ?}} Yes. Contrary to \cite{he2018rethinking}, we believe ImageNet pre-training is still necessary if we can transfer some of the knowledge learned in the ImageNet dataset to downstream tasks.

\textbf{\emph{Why DaR sampling strategy can maintain or improve classification accuracy even when training cost is reduced?}} \textbf{Hard Examples Under-fitting.} Standard cross entropy criterion treats all input examples equally even when they are already correctly classified. As discussed in related work~\cite{lin2017focal}, these easy, correctly classified examples still generate a large proportion of the empirical loss. However, our DaR sampling strategy progressively drop easy examples, which enforce the network to put more focus on hard, mis-classified examples.

\textbf{\emph{Why DaR pre-trained models transfer better?}} \textbf{Task Over-fitting.} Similar to the phenomenon that models over-fit to the training set (models achieve high accuracy on the training set but low accuracy on the never-seen test set), we hypothesis that models also tend to over-fit to the task. By over-fitting to the task, we mean models achieve good generalization on the main tasks they are trained on (\eg pre-training on image classification) but fail to generalize well on other tasks (\eg transferring to downstream tasks). As an ImageNet pre-trained model, the general goal is to learn a good feature representation that may help downstream perception tasks. However this goal is different than the image classification objective, which is to minimize the empirical cross entropy loss of the training data. That is, models that perform extremely well on pre-trained task and may not transfer well to other tasks. To verify our hypothesis, we visualize the ResNet-110~\cite{he2016deep} features before the final classifier (\ie fully connected layer) on the CIFAR-10~\cite{krizhevsky2009cifardataset} test set by projecting the 256-dimension features to 2 dimensions using t-SNE~\cite{maaten2008visualizing} in Figure~\ref{fig:t-sne}. As can be seen from the visualization, training with \iid data (Figure~\ref{fig:t-sne} (a)) results in a more separate clustering than training with non-\iid data (Figure~\ref{fig:t-sne} (b)). From the accuracy prospective, DaR-learned features are enough to distinguish between different classes but the margin is just not as large as using \iid samples. This observation suggests \iid training data might cause models to over-fit the task. The same phenomenon is also observed in~\cite{he2018rethinking}. Task over-fitting can well explain why training from scratch outperforms its pre-trained counterparts (over-fitted on image classification) given enough training iterations.

\textbf{\emph{Is DaR the same as Hard Example Mining?}} \textbf{No.} 
To validate that the DaR is not a simplified hard example mining approach, we compare our DaR with a hard example re-weighting solution. Specifically, based on the loss values of all samples obtained in the previous training epoch, we re-weight the importance of each sample in the next epoch, \ie larger loss values lead to larger training weights. The 5 run results on CIFAR10 with ResNet-110 are 92.09/92.45/92.22/91.52/92.10 (mean: 92.076, std: 0.307), which are worse than that of DaR. We consider that using the ``picking" operation to refresh the knowledge embedded in all training samples is an important step to learn a reliable network.  

\begin{table}[h]\setlength{\tabcolsep}{1pt}
	\centering
    \resizebox{0.4\textwidth}{!}{
    \begin{tabular}{l|c|c|c|c}
	 Method & Many-shot & Medium-shot & Few-shot & Overall\\
    \hline
    Baseline & 47.7 & 13.8 & 0.6 & 25.1\\
	\hline
	DaR (ours) & 43.6 & 19.6 & 2.7 & 26.5 \\
		\end{tabular}
        }
		\caption{Top-1 accuracy on ImageNet-LT ($\%$).}
\label{tab:imagenet_lt}
%  \vspace{-4mm}
\end{table}
\textbf{\emph{Is DaR helpful to real applications?}} \textbf{Yes.}
To validate that DaR is also helpful to train model on non-\iid data, we further perform experiments in long-tail image classification where the data is naturally non-\iid. We train a ResNet-10~\cite{he2016deep} model on the ImageNet-LT~\cite{liu2019large} dataset. The result is shown in Table~\ref{tab:imagenet_lt}, using DaR improves the overall accuracy by $1.4\%$, which suggests it is helpful to real applications.
\section{Conclusion}
We present an efficient pre-training strategy by sampling non-\iid training data named ``Drop-and-Refresh'' (DaR) aiming at reducing training time given limited computational resource. The method is validate on various image classification benchmarks. Specifically, we achieve comparable performance using only 86\% computation on the ImageNet dataset. We further show that the DaR pre-trained models have no loss in generalization ability in many downstream tasks. More interestingly, we observe consistent improvements for these downstream tasks when DaR pre-trained models are used. We will leave it as a future work to study how downstream tasks can benefit from different pre-training policies.

% {\small
% \bibliographystyle{aaai}
% \bibliography{egbib}
% }

{\small
\bibliographystyle{ieee_fullname}
\bibliography{egbib}

\begin{thebibliography}{10}\itemsep=-1pt

\bibitem{cao2017realtime}
Zhe Cao, Tomas Simon, Shih-En Wei, and Yaser Sheikh.
\newblock Realtime multi-person 2d pose estimation using part affinity fields.
\newblock In {\em {CVPR}}, 2017.

\bibitem{mmdetection2018}
Kai Chen, Jiangmiao Pang, Jiaqi Wang, Yu Xiong, Xiaoxiao Li, Shuyang Sun,
  Wansen Feng, Ziwei Liu, Jianping Shi, Wanli Ouyang, Chen~Change Loy, and
  Dahua Lin.
\newblock mmdetection.
\newblock \url{https://github.com/open-mmlab/mmdetection}, 2018.

\bibitem{deeplabv3plus2018}
Liang-Chieh Chen, Yukun Zhu, George Papandreou, Florian Schroff, and Hartwig
  Adam.
\newblock Encoder-decoder with atrous separable convolution for semantic image
  segmentation.
\newblock In {\em ECCV}, 2018.

\bibitem{cheng18decoupled}
Bowen Cheng, Yunchao Wei, Honghui Shi, Rogerio Feris, Jinjun Xiong, and Thomas
  Huang.
\newblock Decoupled classification refinement: Hard false positive suppression
  for object detection.
\newblock {\em arXiv preprint arXiv:1810.04002}, 2018.

\bibitem{cheng2018revisiting}
Bowen Cheng, Yunchao Wei, Honghui Shi, Rogerio Feris, Jinjun Xiong, and Thomas
  Huang.
\newblock Revisiting rcnn: On awakening the classification power of faster
  rcnn.
\newblock In {\em {ECCV}}, 2018.

\bibitem{cordts2016cityscapes}
Marius Cordts, Mohamed Omran, Sebastian Ramos, Timo Rehfeld, Markus Enzweiler,
  Rodrigo Benenson, Uwe Franke, Stefan Roth, and Bernt Schiele.
\newblock The cityscapes dataset for semantic urban scene understanding.
\newblock In {\em {CVPR}}, 2016.

\bibitem{deng2009imagenet}
Jia Deng, Wei Dong, Richard Socher, Li-Jia Li, Kai Li, and Li Fei-Fei.
\newblock Imagenet: A large-scale hierarchical image database.
\newblock In {\em {CVPR}}, 2009.

\bibitem{felzenszwalb2009object}
Pedro~F Felzenszwalb, Ross~B Girshick, David McAllester, and Deva Ramanan.
\newblock Object detection with discriminatively trained part-based models.
\newblock {\em {IEEE TPAMI}}, 2009.

\bibitem{goyal2017accurate}
Priya Goyal, Piotr Doll{\'a}r, Ross Girshick, Pieter Noordhuis, Lukasz
  Wesolowski, Aapo Kyrola, Andrew Tulloch, Yangqing Jia, and Kaiming He.
\newblock Accurate, large minibatch sgd: training imagenet in 1 hour.
\newblock {\em arXiv preprint arXiv:1706.02677}, 2017.

\bibitem{he2018rethinking}
Kaiming He, Ross Girshick, and Piotr Dollár.
\newblock Rethinking imagenet pre-training.
\newblock {\em arXiv preprint arXiv:1811.08883}, 2018.

\bibitem{he2017mask}
Kaiming He, Georgia Gkioxari, Piotr Doll{\'a}r, and Ross Girshick.
\newblock Mask r-cnn.
\newblock In {\em {ICCV}}, 2017.

\bibitem{he2016deep}
Kaiming He, Xiangyu Zhang, Shaoqing Ren, and Jian Sun.
\newblock Deep residual learning for image recognition.
\newblock In {\em {CVPR}}, 2016.

\bibitem{howard2017mobilenets}
Andrew~G Howard, Menglong Zhu, Bo Chen, Dmitry Kalenichenko, Weijun Wang,
  Tobias Weyand, Marco Andreetto, and Hartwig Adam.
\newblock Mobilenets: Efficient convolutional neural networks for mobile vision
  applications.
\newblock {\em arXiv preprint arXiv:1704.04861}, 2017.

\bibitem{huang2017densely}
Gao Huang, Zhuang Liu, Laurens Van Der~Maaten, and Kilian~Q Weinberger.
\newblock Densely connected convolutional networks.
\newblock In {\em {CVPR}}, 2017.

\bibitem{torch2018segment}
Zilong Huang, Yunchao Wei, Xinggang Wang, and Wenyu Liu.
\newblock A pytorch semantic segmentation toolbox, 2018.

\bibitem{jia2018highly}
Xianyan Jia, Shutao Song, Wei He, Yangzihao Wang, Haidong Rong, Feihu Zhou,
  Liqiang Xie, Zhenyu Guo, Yuanzhou Yang, Liwei Yu, et~al.
\newblock Highly scalable deep learning training system with mixed-precision:
  Training imagenet in four minutes.
\newblock {\em arXiv preprint arXiv:1807.11205}, 2018.

\bibitem{joshi2009multi}
Ajay~J Joshi, Fatih Porikli, and Nikolaos Papanikolopoulos.
\newblock Multi-class active learning for image classification.
\newblock In {\em {CVPR}}, 2009.

\bibitem{katharopoulos2018not}
Angelos Katharopoulos and Fran{\c{c}}ois Fleuret.
\newblock Not all samples are created equal: Deep learning with importance
  sampling.
\newblock {\em arXiv preprint arXiv:1803.00942}, 2018.

\bibitem{krizhevsky2009cifardataset}
Alex Krizhevsky.
\newblock Learning multiple layers of features from tiny images.
\newblock Technical report, Citeseer, 2009.

\bibitem{lakshminarayanan2017simple}
Balaji Lakshminarayanan, Alexander Pritzel, and Charles Blundell.
\newblock Simple and scalable predictive uncertainty estimation using deep
  ensembles.
\newblock In {\em {NIPS}}, 2017.

\bibitem{li2018scale}
Jianan Li, Xiaodan Liang, ShengMei Shen, Tingfa Xu, Jiashi Feng, and Shuicheng
  Yan.
\newblock Scale-aware fast r-cnn for pedestrian detection.
\newblock {\em IEEE transactions on Multimedia}, 2018.

\bibitem{liang2016semantic}
Xiaodan Liang, Xiaohui Shen, Jiashi Feng, Liang Lin, and Shuicheng Yan.
\newblock Semantic object parsing with graph lstm.
\newblock In {\em {ECCV}}, 2016.

\bibitem{lin2017refinenet}
Guosheng Lin, Anton Milan, Chunhua Shen, and Ian Reid.
\newblock Refinenet: Multi-path refinement networks for high-resolution
  semantic segmentation.
\newblock In {\em {CVPR}}, 2017.

\bibitem{lin2017feature}
Tsung-Yi Lin, Piotr Doll{\'a}r, Ross Girshick, Kaiming He, Bharath Hariharan,
  and Serge Belongie.
\newblock Feature pyramid networks for object detection.
\newblock In {\em {CVPR}}, 2017.

\bibitem{lin2017focal}
Tsung-Yi Lin, Priya Goyal, Ross Girshick, Kaiming He, and Piotr Dollar.
\newblock Focal loss for dense object detection.
\newblock In {\em {ICCV}}, 2017.

\bibitem{lin2014microsoft}
Tsung-Yi Lin, Michael Maire, Serge Belongie, James Hays, Pietro Perona, Deva
  Ramanan, Piotr Doll{\'a}r, and C~Lawrence Zitnick.
\newblock Microsoft coco: Common objects in context.
\newblock In {\em {ECCV}}, 2014.

\bibitem{liu2019large}
Ziwei Liu, Zhongqi Miao, Xiaohang Zhan, Jiayun Wang, Boqing Gong, and Stella~X
  Yu.
\newblock Large-scale long-tailed recognition in an open world.
\newblock In {\em {CVPR}}, 2019.

\bibitem{loshchilov2015online}
Ilya Loshchilov and Frank Hutter.
\newblock Online batch selection for faster training of neural networks.
\newblock {\em arXiv preprint arXiv:1511.06343}, 2015.

\bibitem{ma2018shufflenetv2}
Ningning Ma, Xiangyu Zhang, Hai-Tao Zheng, and Jian Sun.
\newblock Shufflenet v2: Practical guidelines for efficient cnn architecture
  design.
\newblock In {\em {ECCV}}, 2018.

\bibitem{maaten2008visualizing}
Laurens van~der Maaten and Geoffrey Hinton.
\newblock Visualizing data using t-sne.
\newblock {\em Journal of machine learning research}, 2008.

\bibitem{mahajan2018exploring}
Dhruv Mahajan, Ross Girshick, Vignesh Ramanathan, Kaiming He, Manohar Paluri,
  Yixuan Li, Ashwin Bharambe, and Laurens van~der Maaten.
\newblock Exploring the limits of weakly supervised pretraining.
\newblock In {\em {ECCV}}, 2018.

\bibitem{micikevicius2017mixed}
Paulius Micikevicius, Sharan Narang, Jonah Alben, Gregory Diamos, Erich Elsen,
  David Garcia, Boris Ginsburg, Michael Houston, Oleksii Kuchaev, Ganesh
  Venkatesh, et~al.
\newblock Mixed precision training.
\newblock {\em arXiv preprint arXiv:1710.03740}, 2017.

\bibitem{newell2016stacked}
Alejandro Newell, Kaiyu Yang, and Jia Deng.
\newblock Stacked hourglass networks for human pose estimation.
\newblock In {\em {ECCV}}, 2016.

\bibitem{ogawa2014safe}
Kohei Ogawa, Yoshiki Suzuki, Shinya Suzumura, and Ichiro Takeuchi.
\newblock Safe sample screening for support vector machines.
\newblock {\em arXiv preprint arXiv:1401.6740}, 2014.

\bibitem{osband2016deep}
Ian Osband, Charles Blundell, Alexander Pritzel, and Benjamin Van~Roy.
\newblock Deep exploration via bootstrapped dqn.
\newblock In {\em {NIPS}}, 2016.

\bibitem{paszke2017automatic}
Adam Paszke, Sam Gross, Soumith Chintala, Gregory Chanan, Edward Yang, Zachary
  DeVito, Zeming Lin, Alban Desmaison, Luca Antiga, and Adam Lerer.
\newblock Automatic differentiation in pytorch.
\newblock In {\em NIPS-W}, 2017.

\bibitem{ren2015faster}
Shaoqing Ren, Kaiming He, Ross Girshick, and Jian Sun.
\newblock Faster r-cnn: Towards real-time object detection with region proposal
  networks.
\newblock In {\em {NIPS}}, 2015.

\bibitem{sandler2018mobilenetv2}
Mark Sandler, Andrew Howard, Menglong Zhu, Andrey Zhmoginov, and Liang-Chieh
  Chen.
\newblock Mobilenetv2: Inverted residuals and linear bottlenecks.
\newblock In {\em {CVPR}}, 2018.

\bibitem{schaul2015prioritized}
Tom Schaul, John Quan, Ioannis Antonoglou, and David Silver.
\newblock Prioritized experience replay.
\newblock {\em arXiv preprint arXiv:1511.05952}, 2015.

\bibitem{sergeev2018horovod}
Alexander Sergeev and Mike Del~Balso.
\newblock Horovod: fast and easy distributed deep learning in tensorflow.
\newblock {\em arXiv preprint arXiv:1802.05799}, 2018.

\bibitem{Shen2017DSOD}
Zhiqiang Shen, Zhuang Liu, Jianguo Li, Yu-Gang Jiang, Yurong Chen, and
  Xiangyang Xue.
\newblock Dsod: Learning deeply supervised object detectors from scratch.
\newblock In {\em {ICCV}}, 2017.

\bibitem{shen2017learning}
Zhiqiang Shen, Honghui Shi, Rogerio Feris, Liangliang Cao, Shuicheng Yan, Ding
  Liu, Xinchao Wang, Xiangyang Xue, and Thomas~S Huang.
\newblock Learning object detectors from scratch with gated recurrent feature
  pyramids.
\newblock {\em arXiv preprint arXiv:1712.00886}, 2017.

\bibitem{shen2019improving}
Zhiqiang Shen, Honghui Shi, Jiahui Yu, Hai Phan, Rogerio Feris, Liangliang Cao,
  Ding Liu, Xinchao Wang, Thomas Huang, and Marios Savvides.
\newblock Improving object detection from scratch via gated feature reuse.

\bibitem{shrivastava2016training}
Abhinav Shrivastava, Abhinav Gupta, and Ross Girshick.
\newblock Training region-based object detectors with online hard example
  mining.
\newblock In {\em {CVPR}}, 2016.

\bibitem{sun2017integral}
Xiao Sun, Bin Xiao, Fangyin Wei, Shuang Liang, and Yichen Wei.
\newblock Integral human pose regression.
\newblock In {\em ECCV}, 2018.

\bibitem{szegedy2015going}
Christian Szegedy, Wei Liu, Yangqing Jia, Pierre Sermanet, Scott Reed, Dragomir
  Anguelov, Dumitru Erhan, Vincent Vanhoucke, and Andrew Rabinovich.
\newblock Going deeper with convolutions.
\newblock In {\em {CVPR}}, pages 1--9, 2015.

\bibitem{szegedy2016rethinking}
Christian Szegedy, Vincent Vanhoucke, Sergey Ioffe, Jon Shlens, and Zbigniew
  Wojna.
\newblock Rethinking the inception architecture for computer vision.
\newblock In {\em {CVPR}}, 2016.

\bibitem{wei2016convolutional}
Shih-En Wei, Varun Ramakrishna, Takeo Kanade, and Yaser Sheikh.
\newblock Convolutional pose machines.
\newblock In {\em {CVPR}}, 2016.

\bibitem{wei2018ts2c}
Yunchao Wei, Zhiqiang Shen, Bowen Cheng, Honghui Shi, Jinjun Xiong, Jiashi
  Feng, and Thomas Huang.
\newblock Ts2c: Tight box mining with surrounding segmentation context for
  weakly supervised object detection.
\newblock In {\em {ECCV}}, 2018.

\bibitem{wu2018group}
Yuxin Wu and Kaiming He.
\newblock Group normalization.
\newblock In {\em {ECCV}}, 2018.

\bibitem{xiao2018simple}
Bin Xiao, Haiping Wu, and Yichen Wei.
\newblock Simple baselines for human pose estimation and tracking.
\newblock In {\em ECCV}, 2018.

\bibitem{xie2017aggregated}
Saining Xie, Ross Girshick, Piotr Doll{\'a}r, Zhuowen Tu, and Kaiming He.
\newblock Aggregated residual transformations for deep neural networks.
\newblock In {\em {CVPR}}, 2017.

\bibitem{zhang2017mixup}
Hongyi Zhang, Moustapha Cisse, Yann~N Dauphin, and David Lopez-Paz.
\newblock mixup: Beyond empirical risk minimization.
\newblock {\em arXiv preprint arXiv:1710.09412}, 2017.

\bibitem{zhang2018context}
Hang Zhang, Kristin Dana, Jianping Shi, Zhongyue Zhang, Xiaogang Wang, Ambrish
  Tyagi, and Amit Agrawal.
\newblock Context encoding for semantic segmentation.
\newblock In {\em {CVPR}}, 2018.

\bibitem{zhang2016faster}
Liliang Zhang, Liang Lin, Xiaodan Liang, and Kaiming He.
\newblock Is faster r-cnn doing well for pedestrian detection?
\newblock In {\em {ECCV}}, 2016.

\bibitem{zhao2017pyramid}
Hengshuang Zhao, Jianping Shi, Xiaojuan Qi, Xiaogang Wang, and Jiaya Jia.
\newblock Pyramid scene parsing network.
\newblock In {\em {CVPR}}, 2017.

\bibitem{zhu2018scratchdet}
Rui Zhu, Shifeng Zhang, Xiaobo Wang, Longyin Wen, Hailin Shi, Liefeng Bo, and
  Tao Mei.
\newblock Scratchdet: Exploring to train single-shot object detectors from
  scratch.
\newblock {\em arXiv preprint arXiv:1810.08425}, 2018.

\end{thebibliography}
}

\end{document}